\pdfoutput=1
\documentclass[twoside]{article}
\usepackage[accepted]{aistats2017}

\usepackage[utf8]{inputenc} % allow utf-8 input
\usepackage[T1]{fontenc}    % use 8-bit T1 fonts
\usepackage{url}            % simple URL typesetting
\usepackage{booktabs}       % professional-quality tables
\usepackage{amsfonts}       % blackboard math symbols
\usepackage{nicefrac}       % compact symbols for 1/2, etc.
\usepackage{microtype}      % microtypography
\usepackage{amsmath}
\usepackage{bbm}
\usepackage{algorithmic, algorithm}
\usepackage{amsthm}
\usepackage{verbatim}
\usepackage{graphicx}
\usepackage{caption}
\usepackage[labelformat=simple]{subcaption}
\usepackage[round]{natbib}
\usepackage{xparse}

\newtheoremstyle{sltheorem}
{3pt}% <Space above>
{3pt}% <Space below>
{\slshape}% <Body font>
{}% <Indent amount>
{\bfseries}% <Theorem head font>
{.}% <Punctuation after theorem head>
{.5em}% <Space after theorem headi>
{}% <Theorem head spec (can be left empty, meaning `normal')>
\theoremstyle{sltheorem}

\newtheorem{lemma}{Lemma}
\newtheorem{defn}{Definition}
\newtheorem{example}{Example}
\newtheorem{proposition}{Proposition}

\newcommand{\g}{\,|\,} % given
\newcommand{\te}{\!=\!} % thin equals
\newcommand{\tm}{\!-\!} % thin minus
\newcommand{\ttimes}{\!\times\!} % thin times
\newcommand{\N}{\mathcal{N}}

\newcommand{\E}{\mathbb{E}} 
\newcommand{\targ}{P}
\newcommand{\prop}{Q}
\newcommand{\rb}{\text{rb}}
\newcommand{\chain}[1][]{
X
\ifx\\#1\\
\else
^{(#1)}
\fi
}
\newcommand{\tchain}[1][]{
\tilde{X}
\ifx\\#1\\
\else
^{(#1)}
\fi
}
\newcommand{\estim}[1][]{
Y \ifx\\#1\\
\else
^{(#1)}
\fi
}
\newcommand{\testim}[1][]{
\tilde{Y}
\ifx\\#1\\
\else
^{(#1)}
\fi
}
\newcommand{\pro}{\text{prop}}
\newcommand{\rand}[1][]{
r
\ifx\\#1\\
\else
^{(#1)}
\fi
}
\newcommand{\textif}{\text{ if }}
\newcommand{\tranfunc}[1][]{
\phi
\ifx\\#1\\
\else
^{(#1)}
\fi
}

\begin{document}

\twocolumn[

\aistatstitle{Markov Chain Truncation for Doubly-Intractable Inference}

\aistatsauthor{ Colin Wei \And Iain Murray }

\aistatsaddress{ Stanford University \And University of Edinburgh} ]

\begin{abstract}
Computing \emph{partition functions}, the normalizing constants of probability distributions, is often hard. Variants of importance sampling give unbiased estimates of a normalizer $Z$, however, unbiased estimates of the reciprocal $1/Z$ are harder to obtain. Unbiased estimates of $1/Z$ allow Markov chain Monte Carlo sampling of ``doubly-intractable'' distributions, such as the parameter posterior for Markov Random Fields or Exponential Random Graphs. We demonstrate how to construct unbiased estimates for $1/Z$ given access to black-box importance sampling estimators for $Z$. We adapt recent work on random series truncation and Markov chain coupling, producing estimators with lower variance and a higher percentage of positive estimates than before. Our debiasing algorithms are simple to implement, and have some theoretical and empirical advantages over existing methods.
\end{abstract}

\section{Introduction}

Markov Chain Monte Carlo (MCMC) algorithms can asymptotically draw samples from distributions with intractable normalizing constants. However, sampling from ``doubly-intractable'' distributions \citep{murray2006} is more challenging: direct application of MCMC methods requires the computation of an intractable normalizing constant $Z(\theta)$ at each step (Section~\ref{sec:pmmcmc} has an example). Until recently, the only valid MCMC methods for doubly-intractable distributions required exact samples from distributions with the relevant normalizing constants \citep{moller2006efficient,murray2006}. Drawing exact samples is possible for some high-dimensional distributions \citep{propp1998coupling}, but is hard in general.

\citet{lyne2015russian} provided the first practical and asymptotically correct MCMC method for doubly-intractable distributions that doesn't require exact sampling. This work constructs unbiased estimates of the reciprocal normalizing constants $1/Z(\theta)$ using unbiased estimates of $Z(\theta)$ obtained by importance sampling. A ``Russian roulette'' random series truncation debiases the estimator for $1/Z(\theta)$. The pseudo-marginal framework \citep{andrieu2009pseudo} is then adapted to use these estimates to form an MCMC method.

Inspired by the approach of \citet{glynn2014exact}, we construct unbiased estimates of reciprocal normalizing constants by applying Russian roulette truncations to a Markov chain rather than an importance sampler. Swapping to Markov chains improves two aspects of the estimators, both theoretically and empirically.

First, Russian roulette estimates of the reciprocal normalizer are not guaranteed to be positive. It can be shown that there is no general procedure to construct a strictly positive unbiased estimator by debiasing estimates of the normalizer \citep{jacob2015nonnegative}. However, we find Markov chain-based estimators are positive more often than corresponding importance sampling estimators, and we test the impact of this difference on a doubly-intractable Markov chain empirically.

Second, Russian roulette forms estimates by truncating an infinite series. In the original scheme, each subsequent term in the series was estimated with an exponentially growing number of importance samples, yet it is still hard to prove that the estimator has finite expectation. Our estimator has provably finite expectation, and only requires a number of Monte Carlo samples linear in the length of the truncated series.

\vspace*{-0.05cm} 
\section{Preliminaries}
\vspace*{-0.05cm} 

For the remainder of this paper, we will assume that we are interested in the partition function $Z(\theta)$ of distributions $p(x\g \theta) \te \frac{p^*(x\g \theta)}{Z(\theta)}$, parameterized by $\theta$. Here, $p^*(x\g\theta)$ is the unnormalized probability, defining the partition function $Z(\theta) = \int p^*(x\g\theta)\,\mathrm{d}x$. We will omit the parameters $\theta$ when we only need to consider one normalizing constant.

Importance sampling can give an unbiased estimator of a normalizer $Z$. The method needs a target distribution $\targ(X)\te\targ^*(X)/Z$ that has the normalization constant we are interested in, and a proposal distribution $\prop$ with support on the same state space~$\mathcal{X}$. The unbiased estimator for $Z$ is an average of importance weights, $w(X)\te P^*(X)/Q(X)$, for states sampled from $\prop$:
\begin{align*}
\E_{X \sim \prop} \left[\frac{\targ^*(X)}{\prop(X)}\right] = Z.
\end{align*}
In general, the importance sampling target $\targ$ and our original distribution of interest $p$ do not have to be the same. For example, annealed importance sampling (AIS) \citep{neal2001annealed}, performs importance sampling on an augmented state space.

We will require an unbiased estimate of $1/Z$. Jensen's inequality states that the reciprocal of an importance sampling estimate is biased, and so needs correcting.

\subsection {Russian Roulette Truncation}

Russian roulette truncation can be used to obtain unbiased estimates of $1/Z$. The method was first introduced in the physics literature \citep{1975ptsw.rept.....C, lux1991monte}, while we rely on the formulation presented by \citet{mcleish2011general}, \citet{glynn2014exact} and \citet{lyne2015russian}. For our specific setting, the truncation scheme depends on a sequence of estimators $\estim = (\estim[i] : i \ge 0)$ which satisfy the property that $\lim_{i \rightarrow \infty} \E[\estim[i]] = 1/Z$. The procedure involves drawing a random integer $N$, independent of $Y$, and then taking the sum
\begin {equation}
\label{eq:rroulette}
S = \estim[0] + \sum_{i = 1}^{N} \frac{\estim[i] - \estim[i - 1]}{\Pr(N \ge i)}.
\end {equation}
Provided that our estimators $Y$ are ``good enough'', we will have $\E[S] = 1/Z$. For example, \cite{glynn2014exact} rely on the following lemma to show unbiasedness of their estimators:
\begin{lemma}
\label{lem:finiteexpectation}
$\E[S] = 1/Z$ if the following 
holds:
\begin{equation}
\label{eq:correctestim}
\textstyle 
\E\left[ | \estim[0] | + \sum_{i = 1}^{\infty} | \estim[i] - \estim[i - 1]|\right] < \infty.
\end{equation}
\end{lemma}
The estimator in \eqref{eq:rroulette} is a Monte Carlo estimate of the infinite sum $\estim[0] + \sum_{i = 1}^{\infty} (\estim[i] \!-\! \estim[i - 1])$, which relies on the $Y$ estimates becoming correct asymptotically. Condition \eqref{eq:correctestim} guarantees that the expectation of this Monte Carlo estimate is finite.

We can now define a baseline estimator inspired by \citet{lyne2015russian}. This estimator uses independent samples $\chain[0], \ldots, \chain[N] \sim \prop$ from which we set
\begin{equation}
\label{eq:iae}
\textstyle
\estim[i] = \frac{i + 1}{\sum_{j = 0}^{i} w(\chain[j])}.
\end{equation}
We will refer to this estimator as the Increasing Averages Estimator (IAE)\@. \citet{lyne2015russian} used a similar estimator, but with an exponentially increasing number of samples for each $\estim[i]$. So that we can make direct comparisons of individual design choices, all of the methods that we consider in this paper form estimates $\estim[i]$ based on a number of samples linear in $i$. Our proposed estimators work in this regime, and we could choose the distribution on $N$ without worrying about running time growing out of control. However, our experiments are testing the individual theoretical proposals in this paper, not against the whole system that was originally proposed.

\subsection{Pseudo-Marginal Markov Chain}
\label{sec:pmmcmc}
We now review how to apply these unbiased estimates for $1/Z$ inside a pseudo-marginal outer MCMC loop. Recall that we have a class of densities $p(x\g \theta) = p^*(x\g \theta)/Z(\theta)$. Let $\pi(\theta)$ be a prior over the parameters, and $y$ be a set of observations. Then the target posterior distribution is given by
\begin{align*}
\pi(\theta \g y) \propto \frac{p^*(y \g \theta)\, \pi(\theta)}{Z(\theta)}.
\end{align*}
Standard Metropolis--Hastings sampling of this distribution, with proposal $t(\theta'; \theta)$, computes the term
\begin{align*}
\min \left[1,\; \frac{p^*(y \g \theta')\, \pi(\theta')\, t(\theta ; \theta')\, Z(\theta)}{p^*(y \g \theta)\, \pi(\theta)\, t(\theta' ; \theta)\, Z(\theta')} \right],
\end{align*}
which requires the intractable ratio $Z(\theta)/Z(\theta')$.

A pseudo-marginal transition rule avoids needing to evaluate the normalizers exactly. Following the notation of \citet{murray2016}, let $f(\theta) = p^*(y \g \theta) \pi(\theta)/Z(\theta)$, with an unbiased estimate $\hat{f}$. If $\hat{f}$ is always positive, we can perform Metropolis--Hastings on the augmented state pair $(\theta, \hat{f})$. From the current state pair $(\theta, \hat{f})$, we propose a new state $\theta'$ with estimate $\hat{f'}$ and accept with probability
\begin{align*}
    \textstyle 
\min \left[1, \frac{\hat{f'}}{\hat{f}} \frac{ t(\theta ; \theta')}{t(\theta' ; \theta)}\right].
\end{align*}
Unfortunately, the roulette estimator \eqref{eq:rroulette} can be negative if $\estim[i] \tm \estim[i - 1] < 0$ for many values of~$i$. \citet{lyne2015russian} provide a clever way to avoid this ``sign problem'': replace the acceptance probability with
\begin{align*}
    \textstyle 
\min \left[1, \frac{|\hat{f'}|}{|\hat{f}|} \frac{ t(\theta ; \theta')}{t(\theta' ; \theta)}\right].
\end{align*}
Then for each visited state $(\theta_i, \hat{f}_i)$, save $\sigma_i$, the sign of $\hat{f}_i$ such that $\hat{f}_i = \sigma_i | \hat{f}_i|$. Finally, when estimating the expectation of some function $h(\theta)$ over the posterior, the approximation $\sum_i h(\theta_i) \sigma_i / \sum_i \sigma_i$ is a consistent estimator for $\E_{\pi(\theta \g y)} [h(\theta)]$.

A drawback to pseudo-marginal methods is that high variability in the estimator $\hat{f}(\theta)$ can encourage ``sticking'', as the same estimate $\hat{f}$ must be kept until a new state $\theta'$ is accepted. Furthermore, although the sign-normalized estimators are consistent, they will have high variance if a large fraction of the signs are negative. The construction of our Markov chain based estimators is motivated by the desire to address these issues.

\section{Using a Markov Chain to Debias Importance Sampling Estimates}
As motivation, we observe that the expectation of the inverse importance weights with respect to $\targ$ is $1/Z$:
\begin{align*}
    \textstyle 
    \E_{X \sim \targ}\left[\frac{1}{w(X)}\right] = \int \frac{\targ(X) \prop(X)}{\targ^*(X)}\;\mathrm{d}X = \frac{1}{Z}.
\end{align*}
Thus, samples drawn from $\targ$ can provide unbiased estimates of $1/Z$. Although we can sample some target distributions $\targ$ using coupling from the past \citep{propp1998coupling}, for many choices of $\targ$ no tractable exact sampling algorithm is known. However, using the tools from the previous section, we actually only need a sequence of samples whose distributions converge to $\targ$. We can obtain these with Markov chain Monte Carlo methods. We use the Metropolis--Hastings algorithm with proposals $\prop$ taken from an importance sampler.

We will use $(\chain = \chain[i] : i \ge 0)$ to denote the states of our Markov chain. We can run a Markov chain whose stationary distribution converges to $\targ$ as follows:

\vspace*{-0.3cm}
\begin{enumerate}
\item At time step $i$, draw a new state $\chain[i + 1]_{\pro} \sim \prop$.
\item Compute the acceptance ratio
\begin{align*}
    \textstyle 
a &= \textstyle \min \left[1, \frac{\targ(\chain[i + 1]_{\pro})\, \prop(\chain[i])}{\prop(\chain[i + 1]_{\pro})\, \targ(\chain[i])}\right]
=\min \left[1, \frac{w(\chain[i + 1]_{\pro})}{w(\chain[i])}\right]\!.
\end{align*}
\item Draw a uniform random value $\rand[i] \in [0, 1]$ and set\\[-0.2in] 
\begin{align*}
    \textstyle 
\chain[i + 1] = \begin{cases}
\chain[i] &\textif \rand[i] < a\\
\chain[i + 1]_{\pro} &\textif \rand[i] \ge a.
\end{cases}
\end{align*}
\end{enumerate}

\vspace*{-0.2cm}
This chain and associated weights $w(X)\te P^*(X)/Q(X)$ forms the backbone of our proposed debiasing schemes.

We need an asymptotically correct estimate $\estim$. One obvious choice is $\estim[i] = 1/w(\chain[i])$, where
\begin{align*}
\lim_{i \rightarrow \infty} \E[\estim[i]] = \lim_{i \rightarrow \infty} \E[1/w(\chain[i])] = 1/Z,
\end{align*}
since the distribution of $\chain[i]$ approaches $\targ$. The main problem with this choice is that $\E[|\estim[i] - \estim[i - 1]|]$ does not decay, as the chain might make large jumps from $\chain[i - 1]$ to $\chain[i]$. As a result, the variance of the Russian roulette truncations will be high, and the final estimator might even have infinite expectation.

\citet{glynn2014exact} suggests instead finding two sequences: $\estim = (\estim[i] : i \ge 0)$ and $\testim = (\testim[i] : i \ge 0)$ such that $\testim[i]$ follows the same distribution as $\estim[i]$, but $\estim[i]$ and $\testim[i - 1]$ are likely to ``couple'' together.~Then\\[-0.2in]
\begin{align}
    \textstyle 
\label{eq:estimtilde}
S = \estim[0] + \sum_{i = 1}^N \frac{\estim[i] - \testim[i - 1]}{\Pr(N \ge i)}\\[-0.3in] \notag
\end{align}
is an unbiased estimator of $1/Z$, since $\estim[i]$ and $\testim[i]$ follow the same distribution.

The pair of estimators $\estim[i]$ and $\testim[i]$ are constructed from Markov chains that share random numbers.
Our Markov chain $\chain = (\chain[i] : i \ge 0)$ uses a transition rule $\tranfunc \colon \mathcal{X} \times \mathcal{X} \times [0, 1] \rightarrow \mathcal{X}$, which uses a random number $\rand$ to make each accept/reject decision:\\[-0.1in]
\begin{equation*}
\begin{split}
\chain[i + 1] &= \tranfunc(\chain[i], \chain[i + 1]_{\pro}, \rand[i]),\\[-0.05in]
\end{split}
\end{equation*}
where $\tranfunc$ returns either the previous or proposed state according to the Metropolis--Hastings rule.

In what follows we write $\tranfunc[i + 1](\cdot) = \tranfunc(\cdot, \chain[i + 1]_{\pro}, \rand[i])$ as the transition function determined by random choices of $\chain[i + 1]_{\pro}$ and $\rand[i]$. We also use $\tchain = (\tchain[i] : i \ge 0)$ to denote a coupled copy of our chain. We would like to describe a coupling between $\chain$ and $\tchain$ so that $\estim[i] = 1/w(\chain[i])$ and $\testim[i] = 1/w(\tchain[i])$ has the desired properties. We investigate alternative couplings in the following sections and defer formal guarantees of finite expectation to Section \ref{sec:finite}.

\vspace*{-0.1cm} 
\subsection{Forward Coupling}
\vspace*{-0.1cm} 

We use the following construction \citep{glynn2014exact}:\\[-0.05in]
\begin{equation}
\begin{split}
\chain[i] &= \tranfunc[i](\tranfunc[i - 1](\ldots(\tranfunc[1](\chain[0])))) \\
\tchain[i] &= \tranfunc[i + 1](\tranfunc[i](\ldots(\tranfunc[2](\chain[0]))))\\[-0.02in]
\end{split} \label{eq:tchainforward}
\end{equation}
The chains $\tchain[i]$ and $\chain[i]$ are dependent, and marginally come from the same distribution. Using $\estim[i] = 1/w(\chain[i])$ and $\testim[i] = 1/w(\tchain[i])$ in \eqref{eq:estimtilde} gives an unbiased estimate for $1/Z$. We can in fact compute $\estim[i]$ and $\testim[i]$ only knowing the sequence of proposed weights $(w(X^{(i)}_{\pro}) : 0 \le i \le N)$ without requiring exact knowledge of the states $\tchain[i]$. This makes it simple to implement our debiasing scheme given access to a black-box importance sampler. We refer to this estimator as the forward coupled estimator (FCE) and illustrate it concretely in Algorithm~\ref{alg:forward}.

\begin{algorithm}
    \small\raggedright
\caption{Forward Coupled Estimator}
\label{alg:forward}
\begin{algorithmic}[1]
  \REQUIRE Target distribution $\targ$ and proposal distribution $\prop$.
  \ENSURE $S$, an unbiased estimate for $1/Z$
  \STATE Draw random stopping time $N$.
  \STATE Draw $\chain[0]_{\pro}, \ldots, \chain[N]_{\pro} \sim \prop$ and initialize $w^{(0)}, \ldots, w^{(N)}$ with $w^{(i)} = w(\chain[i]_{\pro})$.
  \STATE Initialize $S = 1/w^{(0)}$, $w = w^{(0)}$, $\tilde{w} = w^{(0)}$.
  \FOR{$i = 1$ to $N$}
  \STATE Draw $\rand[i - 1] \sim \text{Uniform[0, 1]}$.
  \STATE Compute $a = \min \{1, w^{(i)}/w\}$.
  \STATE Compute $\tilde{a} = \min \{1, w^{(i)}/\tilde{w}\}$.
  \IF{$\rand[i - 1] < a$}
  \STATE Update $w = w^{(i)}$.
  \ENDIF
  \IF{$\rand[i - 1] < \tilde{a}$ and $i > 1$}
  \STATE Update $\tilde{w} = w^{(i)}$.
  \ENDIF
  \STATE Update $S = S + \frac{w^{-1} - \tilde{w}^{-1}}{\Pr(N \ge i)}$.
  \ENDFOR
\end{algorithmic}
\end{algorithm}

The key feature of our estimator is that it attempts to couple together $\chain[i]$ and $\tchain[i - 1]$ by subjecting both chains to the same sequence of random transitions following $\tranfunc[1]$, which is applied to $\chain$ but not $\tchain$. If $\chain[i - 1]$ and $\tchain[i - 2]$ both accept $\chain[i]_{\pro}$ when subjected to $\tranfunc[i]$, then $\chain[i] = \tchain[i - 1]$, and $\chain$ and $\tchain$ couple together. All the subsequent correction terms cancel out if $\chain[i]$ and $\tchain[i - 1]$ have coupled: in \eqref{eq:estimtilde}, $\estim[j] - \testim[j - 1] = 0$ for all $i \le j \le N$. This cancellation serves as a form of variance reduction for our estimator.

We can provide a simple lower bound on the probability of coupling by time step $i$, which also translates into a method for guaranteeing positive estimates.
\begin{lemma}
\label{lem:fcecouple}
For the FCE, if $i \ge 2$,
\begin{align*}
\Pr[\chain[i] \text{ and } \tchain[i - 1] \text{ have coupled }] \ge 1 - \textstyle \frac{2}{i + 1}.
\end{align*}
\end{lemma}

\vspace*{-0.5cm}
\begin{proof}
Let $j^*$ be the smallest $2 \le j \le i$ such that $w(\chain[j]_{\pro}) \ge \max\{w(\chain[0]), w(\chain[1]_{\pro})\}$, if such a $j$ exists. Then both chains $\chain$ and $\tchain$ must accept the proposal at $\tranfunc[j^*]$ since $w(\chain[j^*]_{\pro}) > \max\{w(\chain[j^* - 1]), w(\tchain[j^* - 2])\}$ so the acceptance ratios evaluate to 1. The probability of $j^*$ existing is at least $\frac{i - 1}{i + 1}$, the probability that the largest importance weight is proposed between the second and $i$-th proposal since our importance weights are drawn i.i.d. Thus, the two chains would have coupled with probability at least $1 - 2/(i + 1)$.
\end{proof}

\vspace*{-0.3cm}
If the Markov chain estimator discarded an initial ``burn-in'' period of $T$ time steps, we can guarantee that our estimates will have a probability of at least $1 - 1/T$ of being positive after debiasing. Concretely, define $\estim$ and $\testim$ alternatively so that $\estim[i] = 1/w(\chain[i + T])$ and $\testim[i] = 1/w(\tchain[i + T])$. Then Lemma \ref{lem:fcecouple} implies the following result:
\begin{proposition}
Compute $S$ as in Algorithm \ref{alg:forward}, except allowing for the burn-in of $T$ steps. Then
\begin{align*}
    \textstyle
\Pr[S \ge 0] \ge 1 - \frac{2}{T + 1}
\end{align*}
\end{proposition}

\vspace*{-0.3cm}
This result follows simply from noting that if $\chain[T]$ and $\tchain[T - 1]$ are coupled, then $\estim[i] - \testim[i - 1] = 0$ for $i \ge 1$. In fact, a simple argument can improve the probability to $1 - 1/(T + 1)$, which we omit for space reasons. Our experiments did not use a burn-in period for ease of comparison. Even without burn-in, FCE gives a higher percentage of positive estimates than other formulations.

FCE can have high variance when the underlying importance sampler is variable. If $\chain[1]_{\pro}$ is very large, coupling may be impeded because $\tchain$ does not encounter this proposal, and $\chain$ will have difficulty moving away from $\chain[1]_{\pro}$ due to low acceptance probabilities. Our next estimator improves this situation, although provides fewer guarantees on positive estimates.

\vspace*{-0.1cm} 
\subsection{Backward Coupling}
\vspace*{-0.15cm} 

We use an alternative construction for $\chain$ and $\tchain$, also from \cite{glynn2014exact}:
\begin{align}
\chain[i] &= \tranfunc[N](\tranfunc[N - 1](\ldots(\tranfunc[N - i + 1](\chain[N - i]_{\pro})))),
\label{eq:chainbackward}
\end{align}
where $N$ is the random stopping time of the Russian roulette truncation. We refer to this coupling as ``backwards'' because we process the proposals in reverse. For this estimator, we will simply let $\tchain[i] = \chain[i]$.

We will also reduce the variance of our estimates by computing the expectations of $1/w(\chain[i])$ over the random draws $\rand$ used to determine acceptance. The process of averaging out $\rand$ is a case of a general technique called Rao--Blackwellization, which has been shown to reduce variance when applied to Metropolis--Hastings sampling updates \citep{casella1996}. We can formally express $\estim[i]$ as follows: first independently sample proposals $\chain[0]_{\pro}, \ldots, \chain[N]_{\pro} \sim \prop$. Then
\begin{equation}
    \textstyle
\label{eq:rbbceestim}
\estim[i] = \E \left[ \frac{1}{w(\tranfunc[N](\ldots(\chain[N - i]_{\pro})))} | \chain[N]_{\pro}, \ldots, \chain[N - i]_{\pro}\right]\llap{.}
\end{equation}
Since $\chain[N]_{\pro}, \ldots, \chain[N - i]_{\pro}$ are given, this equation denotes the expectation of $1/w(\chain[i])$ with $\rand[N - 1], \ldots, \rand[N - i]$ averaged out. By the law of iterated expectations, we still have $\lim_{i \rightarrow \infty} \E[\estim[i]] = 1/Z$, so our Rao--Blackwellized estimator is unbiased in $1/Z$.

\begin{example}
In the case where $i = 1$,\\[-0.2in]
\begin{align*}
\estim[1] =& \textstyle \frac{1}{w(\chain[N - 1]_{\pro})} \left( 1- \min \left[1, \frac{w(\chain[N]_{\pro})}{w(\chain[N - 1]_{\pro})} \right]\right) + \\
& \textstyle \frac{1}{w(\chain[N]_{\pro})}  \min \left[1, \frac{w(\chain[N]_{\pro})}{w(\chain[N - 1]_{\pro})} \right]
\end{align*}
\end{example}

\vspace*{-0.1in}
We outline the Rao--Blackwellization process in Algorithm \ref{alg:backwards}. We refer to our estimator as the Rao--Blackwellized backward coupled estimator (RBBCE)\@. Like FCE, RBBCE only requires knowledge of the importance weights, not the states, to run. The algorithm is simple to implement and provably fast in expectation. The following proposition shows that we can perform Rao--Blackwellization essentially ``for free'' on the backward coupled estimator.

\begin{algorithm}[t]
    \small\raggedright
\caption{Rao--Blackwellized Backward Coupled Estimator}
\label{alg:backwards}
\begin{algorithmic}[1]
  \REQUIRE Target distribution $\targ$ and proposal distribution $\prop$.
  \ENSURE $S$, an unbiased estimate for $1/Z$
  \STATE Draw random stopping time $N$.
  \STATE Draw $\chain[0]_{\pro}, \ldots, \chain[N]_{\pro} \sim \prop$ Initialize $w^{(0)}, \ldots, w^{(N)}$ with $w^{(i)} = w(\chain[i]_{\pro})$.
  \STATE Initialize $S = 1/w^{(N)}$, $\estim[0]_{\rb} = 1/w^{(N)}$.
  \FOR{$i = 1$ to $N$} \label{line:outerfor}
  \STATE Find $k$, $0 \le k < i$ such that $w^{(N - k)} = \max_{0 \le j < i} w^{(N - j)}$.
  \IF{$w^{(N - i)} < w^{(N - k)}$}
  \STATE Set $\estim[i]_{\rb} = \estim[k]_{\rb}$.
  \ELSE
  \STATE Initialize $\estim[i]_{\rb} = 0$, $\gamma = 1$.
  \FOR{$j = 0$ to $i - 1$}
  \STATE Update\\[-4ex]
  \begin{align*}
  \estim[i]_{\rb} = \estim[i]_{\rb} + \frac{w^{(N - i + j + 1)}}{w^{(N - i)}} \cdot \gamma \cdot \estim[i - j - 1]_{\rb}
  \end{align*}
  \STATE Update\\[-4ex]
  \begin{align*}
  \gamma = \gamma \cdot \left(1 - \frac{w^{(N - i + j + 1)}}{w^{(N - i)}}\right)
  \end{align*}
  \ENDFOR
  \STATE Update $\estim[i]_{\rb} = \estim[i]_{\rb} + \gamma \cdot \frac{1}{w^{(N - i)}}$.
  \ENDIF
  \STATE Update $S = S + \frac{\estim[i]_{\rb} - \estim[i - 1]_{\rb}}{\Pr(N \ge i)}$.
  \ENDFOR
\end{algorithmic}
\end{algorithm}

\begin{proposition}
\label{prop:rbtime}
Algorithm \ref{alg:backwards} takes expected $O(N)$ running time.
\end{proposition}

\vspace*{-0.5cm}
\begin{proof}
    Following the notation in Algorithm \ref{alg:backwards}, we will let $w^{(i)} = w(\chain[i]_{\pro})$. We can compute the expected runtime of each iteration of the loop at line \ref{line:outerfor}. If the current proposed weight at iteration $i$, $w^{(N - i)}$, is less than $\max_{0 \le j < i} w^{(N - j)}$, then the chain will always accept at this maximum because the acceptance ratio will be 1. In this case, we take $O(1)$ time to update \smash{$\estim[i]_{\rb}$}. If the current proposed weight is greater than $\max_{0 \le j < i} w^{(N - j)}$, then we take $O(i)$ time to compute
    \smash{$Y_{\rb}\kern-2pt{}^{(i)}$}. 
    The probability of this happening is $\frac{1}{i + 1}$ because $w^{(N - i)}, \ldots, w^{(N)}$ are i.i.d.\ draws, so the total expected runtime of each iteration is $O(1) + O(i/(i + 1)) = O(1)$. The loop runs $N$ times giving expected $O(N)$ runtime.\looseness=-1
\end{proof}

\vspace*{-0.4cm}
The $O(N)$ expected time means Rao--Blackwellization only adds a constant cost to the computation of each importance weight, which will be negligible for expensive, low-variance weights. 
We can explain 
``coupling'' in RBBCE as follows: $\estim[i] \!=\! \estim[i - 1]$ unless $w(\chain[N - i]_{\pro}) > \max_{N -i + 1 \le j \le N} w(\chain[j]_{\pro})$, because otherwise the acceptance probability will be~1. Thus, $\estim[i] - \estim[i - 1]$ only has probability $1/(i + 1)$ of being nonzero. In comparison, the difference terms in IAE contribute to higher variance because they are only nonzero if\\[-0.2in]
\begin{align*}
    \textstyle
w(\chain[i]) = \left[\sum_{j = 0}^{i - 1} w(\chain[j])\right] \Big/\, i,\\[-0.3in]
\end{align*}
which occurs with extremely low probability. We find empirically that RBBCE obtains lower variance.

\subsection{Averaging batches of importance weights}
\label{sec:avgscheme}
\vspace*{-0.2cm}

Taking the reciprocal of importance weights in
Algorithm~\ref{alg:forward} or~\ref{alg:backwards} will give high variance
estimates if the weights are occasionally small.
We reduce the variance of the importance weights by averaging over a batch:
\begin{align}
w = \frac{1}{m} \sum_{i = 1}^m \frac{\targ^*(X_i)}{\prop(X_i)}.
\label{eqn:batchw}
\end{align}
One way to justify using these average weights in the Markov chains is
to define new targets and proposals $\targ_m$ and $\prop_m$ on the augmented state space $\mathcal{X}^m$:
\begin{align*}
\prop_m(X_1, \ldots, X_m) &= \textstyle\prod_{i = 1}^{m} \prop(X_i),\\
\targ^*_m(X_1, \ldots, X_m) &= \textstyle\frac{1}{m} \sum_{i = 1}^{m} \targ^*(X_i) \prod_{j \ne i} \prop(X_j).
\end{align*}
Because $\prop$ is normalized, it follows that the normalizer for $\targ^*_m$ is $Z$.
Using $\targ_m$ and $\prop_m$ as target and proposal distributions means that
the weights in Algorithms~\ref{alg:forward} and~\ref{alg:backwards}
become the average of a batch of weights~\eqref{eqn:batchw}.

\section{Unbiasedness in $1/Z$}
\label{sec:finite}
\vspace*{-0.2cm}

Now we will formally establish the unbiasedness properties of our proposed estimators. First, we will formally define when the expectation of a random variable is finite. Our motivation is to characterize when the Law of Large Numbers (LLN) holds for FCE and RBBCE\@.

\begin{defn}
Let $A$ be a random variable with state space $\mathcal{A}$. Let $f \colon \mathcal{A} \rightarrow \mathbb{R}$ be a real-valued function, and let $\lambda$ be the distribution of $A$ on $\mathcal{A}$. Then we say that $A$ has finite expectation if
$
\int_{A} |f(A)| d\lambda < \infty
$
and $A$ has infinite expectation otherwise.
\end{defn}
The distinction between finite and infinite expectation is important because the LLN only applies to random variables with finite expectation. We rely on Lemma \ref{lem:finiteexpectation} to show that FCE has finite expectation whenever $\mathcal{X}$ is a finite state space. We also show that RBBCE always has finite expectation for any choice of state space $\mathcal{X}$.

\begin{proposition}
Let $\prop$ and $\targ$ have full support over $\mathcal{X}$. So long as $\mathcal{X}$ is finite, the output of Algorithm \ref{alg:forward} will have finite expectation and so is unbiased in $1/Z$.
\end{proposition}

\vspace*{-0.5cm}
\begin{proof}
Since $\mathcal{X}$ is finite and $\prop$ and $\targ$ have full support, we can define the maximum and minimum possible importance weights by
\begin{align*}
w_{\min} = \min_{X \in \mathcal{X}} w(X), w_{\max} = \max_{X \in \mathcal{X}} w(X)
\end{align*}
Define $X_{\min}$ and $X_{\max}$ as states corresponding to $w_{\min}$ and $w_{\max}$. Now recall that $\estim[i] = 1/w(\chain[i])$ and $\testim[i - 1] = 1/w(\tchain[i - 1])$. If $X_{\max}$ is proposed by $\tranfunc[j]$ for $2 \le j \le i$, then both $\chain$ and $\tchain$ must accept at $\tranfunc[j]$ with probability 1. In this case, $\estim[i] - \testim[i - 1] = 0$. Now if this does not happen, then the trivial upper bound
\begin{align*}
    \textstyle
|\estim[i] - \testim[i - 1]| \le \left |\frac{1}{w_{\min}} - \frac{1}{w_{\max}}\right|
\end{align*}
must apply. We can upper bound the probability that $X_{\max}$ is not proposed by $(1 - \prop(X_{\max}))^{i - 1}$ since proposals are drawn independently. For $i \ge 2$, this gives us an expected value bound
\begin{align*}
\textstyle
\E[| \estim[i] - \testim[i - 1] |] \le (1 - \prop(X_{\max}))^{i - 1} \left |\frac{1}{w_{\min}} - \frac{1}{w_{\max}}\right|
\end{align*}

\vspace*{-0.5cm}
and therefore
\begin{align*}
&\textstyle \E \left[ |\estim[0]| + \sum_{i = 1}^{\infty} |\estim[i] - \testim[i - 1]|\right] \le \\
&\qquad \E[|\estim[0]|] +  \E[|\estim[1] - \testim[0]|] ~+~ \\
&\textstyle\qquad\sum_{i = 2}^{\infty}(1 - \prop(X_{\max}))^{i - 1} \left |\frac{1}{w_{\min}} - \frac{1}{w_{\max}}\right| < \infty,
\end{align*}
because $(1 - \prop(X_{\max})) < 1$ since $\prop$ has full support on $\mathcal{X}$, and therefore the equation is a geometric series. Thus, \eqref{eq:correctestim} is satisfied (we note that this is for $\estim[i] - \testim[i - 1]$ instead of $\estim[i] - \estim[i - 1]$, but Lemma \ref{lem:finiteexpectation} still applies) so Lemma \ref{lem:finiteexpectation} completes the proof.
\end{proof}
For RBBCE, we can provide even stronger guarantees for unbiasedness. In particular, even if $\mathcal{X}$ is infinite, RBBCE will always have finite expectation so long as $\prop$ and $\targ$ have full support over $\mathcal{X}$ and
\begin{equation}
\label{eq:finiteproptargratio}
\E_{X \sim \prop}[\prop(X)/\targ^*(X)] < \infty.
\end{equation}
This assumption ensures that $\E[\estim[0]] < \infty$ and is a natural assumption to make for reasonable choices of $\targ$. To prove our result, we require the following observation about $\estim[i]$:

\begin{figure*}
    \vspace*{-0.15cm}
\centering 
\includegraphics[scale=0.25]{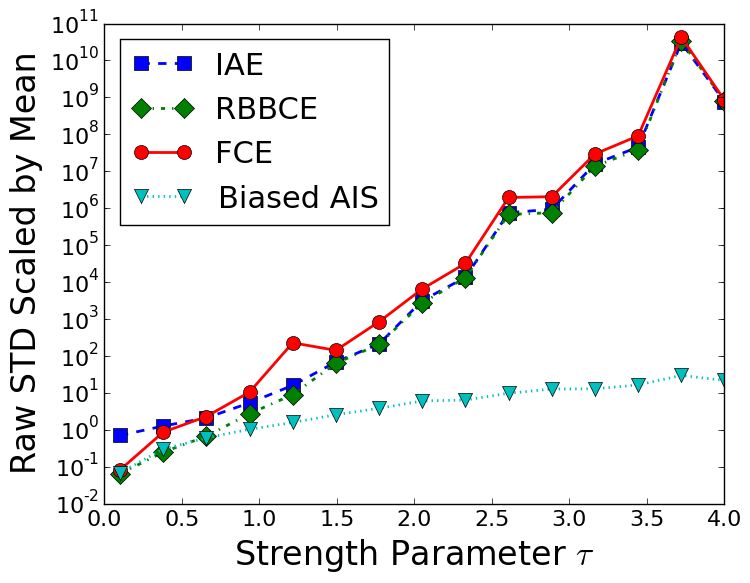}
~~
\includegraphics[scale=0.25]{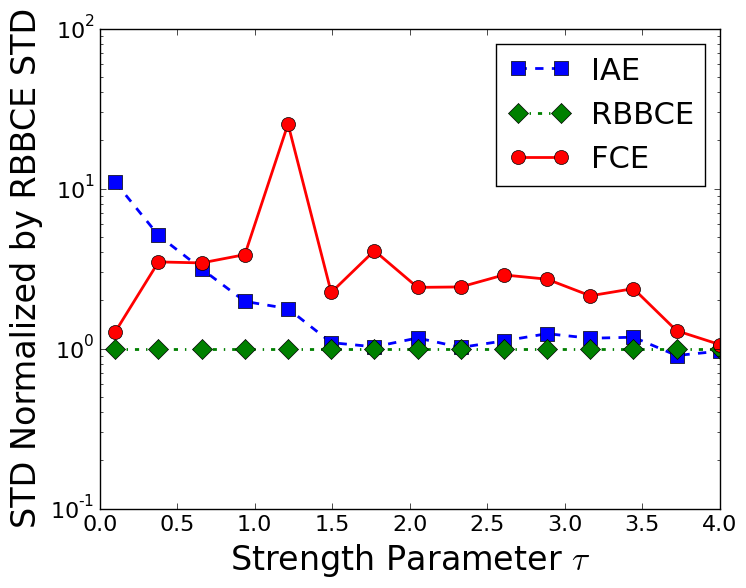}
~~
\includegraphics[scale=0.25]{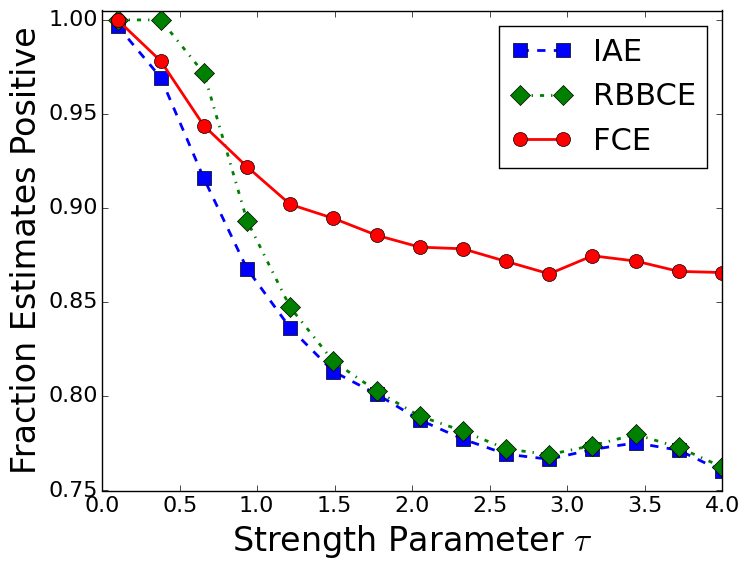}
\vspace*{-0.1cm}
\caption{$1/Z$ estimator performance for Ising models with different values of $\tau$. Each estimator is run for 10,000 trials.
\textbf{Left:}
Standard deviation divided by the mean of the estimator. For IAE, RBBCE, and FCE, this is $1/Z$, which we know exactly. The biased AIS estimator is the inverse importance weights, and we plot empirical standard deviation over empirical mean.
\textbf{Center:}
Each standard deviation is divided by the RBBCE standard deviation, for clearer comparison.
\textbf{Right:}
The fraction of positive estimates returned by each estimator.}
\label{fig:varpos}
\vspace*{-0.2cm}
\end{figure*}

\begin{lemma}
\label{lem:rbbceorder}
Recall that for RBBCE, $\estim[i]$ is defined in \eqref{eq:rbbceestim}. For any $i \ge 1$, $\estim[i] \le \estim[i - 1]$.
\end{lemma}

\vspace*{-0.5cm}
\begin{proof}
For $\estim[i]$, recall that the Markov chain first starts at state $\chain[N - i]_{\pro}$. We will first analyze what happens for each fixed choice of $\tranfunc[N], \ldots, \tranfunc[N - i]$ and then average out the random draws $\rand[N - 1], \ldots, \rand[N - i]$. First, let
\begin{align*}
\hat{\estim[i]} = 1 / w(\tranfunc[N](\ldots(\chain[N - i]_{\pro}))).
\end{align*}
$\hat{\estim[i]}$ denotes an instantiation of $\estim[i]$ without Rao--Blackwellization over random acceptances. Let $k = \max_{0 \le j < i }w(\chain[N - j]_{\pro})$. There are two cases for $\chain[N - i]_{\pro}$: if $w(\chain[N - i]_{\pro}) \le w(\chain[N - k]_{\pro})$, then both the chains for $\hat{\estim[i]}$ and $\hat{\estim[i - 1]}$ must accept $\chain[N - k]_{\pro}$, in which case $\hat{\estim[i]} = \hat{\estim[i - 1]}$. When $w(\chain[N - i]_{\pro}) > w(\chain[N - k]_{\pro})$, consider the first acceptance by the chain for $\hat{\estim[i]}$. Since $w(\chain[N - i]_{\pro}) > w(\chain[N - k]_{\pro})$, the corresponding acceptance ratio for $\hat{\estim[i - 1]}$ is greater than the acceptance ratio for $\hat{\estim[i]}$ at this point. Thus, the chain for $\hat{\estim[i - 1]}$ must also accept at this point, resulting in coupling, so $\hat{\estim[i]} = \hat{\estim[i - 1]}$ again. Finally, if the chain for $\hat{\estim[i]}$ never accepts, then
\begin{align*}
\hat{\estim[i]} = \frac{1}{w(\chain[N - i]_{\pro})} < \frac{1}{w(\chain[N - k]_{\pro})} \le \hat{\estim[i - 1]}.
\end{align*}
In all cases, $\hat{\estim[i]} \le \hat{\estim[i - 1]}$. Thus,
\begin{align*}
\estim[i] - \estim[i - 1] &= \E[\hat{\estim[i]} - \hat{\estim[i - 1]} | \chain[N]_{\pro}, \ldots, \chain[N - i]_{\pro}]\\
& \le 0
\end{align*}
\vspace*{-1.2cm}

\end{proof}
\begin{proposition}
As long as $\prop$ and $\targ$ have full support over $\mathcal{X}$ and \eqref{eq:finiteproptargratio} holds, the output of Algorithm \ref{alg:backwards} will be unbiased in $1/Z$ and have finite expectation.
\end{proposition}

\vspace*{-0.5cm}
\begin{proof}
From Lemma \ref{lem:rbbceorder}, it follows that $\estim[i] - \estim[i - 1] \le 0$ so $\E[|\estim[i] - \estim[i - 1]|] = -\E[\estim[i] - \estim[i - 1]]$ for $i \ge 1$. Therefore,
\begin{align*}
\lim_{n \rightarrow \infty} \sum_{i = 1}^{n} \E[|\estim[i] \!-\! \estim[i - 1]|] &=
- \lim_{n \rightarrow \infty} \sum_{i = 1}^{n} \E[\estim[i] \!-\! \estim[i - 1]]\\ &=
- \lim_{n \rightarrow \infty} (\E[\estim[n]] \!-\! \E[\estim[0]])\\ &= -1/Z + \E[\estim[0]] <  \infty.
\end{align*}
It follows that \eqref{eq:correctestim} holds, so Lemma \ref{lem:finiteexpectation} implies that Algorithm \ref{alg:backwards} provides an output unbiased in $1/Z$.
\end{proof}
\subsection{Comparison To Existing Russian Roulette Estimator}
We do not know of any proofs of finite expectation for the IAE estimator described in \eqref{eq:iae}. A simple example shows that \eqref{eq:correctestim} can be violated.

\begin{example}
Consider the case where $\mathcal{X} = \{0, 1\}$, $\prop(0) = \prop(1) = 1/2$, and $\targ^*(0) = 1$, $\targ^*(1) = 2$. Then if we define $\estim[i]$ as in \eqref{eq:iae},
\begin{align*}
\E\left[| \estim[0] | + \sum_{i = 1}^{\infty} | \estim[i] - \estim[i - 1]| \right]
\end{align*}
is infinite. In particular, \eqref{eq:correctestim} is not satisfied.
\end{example}

\vspace*{-0.5cm}
\begin{proof}[Explanation for claim]
We show that $\E[|\estim[i] - \estim[i - 1]|] = \Omega(1/i)$. Consider the event $E_i$ where at least half of the proposed states $X_j$ for $j < i$ are $0$, and $X_i = 1$. Let $S_i = \sum_{j = 0}^{i} w(X_i)$.
\begin{align*}
    \textstyle
\estim[i] - \estim[i - 1] &= \frac{i + 1}{S_i} - \frac{i}{S_i - 4}
= \frac{S_i - 4i - 4}{S_i(S_i - 4)}
\end{align*}
Now since over half the states $X_j$ with $j < i$ are $0$, it follows that $S_i \le 3i + 4$. Thus, 
\begin{align*}
\estim[i] - \estim[i - 1] \le -\frac{1}{9i + 12}.
\end{align*}
So with probability at least $\Pr[E_i]$, $|\estim[i] - \estim[i - 1]| \ge \frac{1}{9i + 12}$. From inspecting $\prop$, it is evident that $\Pr[E_i] \ge 1/4$, so $\E[|\estim[i] - \estim[i - 1]|] \ge \frac{1}{36i + 48}$. Summing over all $i$ gives a divergent infinite sum.
\end{proof}

Our analysis highlights an advantage of Markov chain based estimators: without the need for a case-by-case analysis or for tuning $\estim[i]$ to require a superlinear number of samples in $i$, our estimators are guaranteed to have the correct expectation for many choices of $\mathcal{X}$ (all choices in the case of RBBCE)\@.

\vspace*{-0.1cm}
\section{Demonstrations}
\vspace*{-0.2cm}

We test empirically how the estimators work in practice. Following \citet{moller2006efficient},
we test our algorithms on a grid Ising model, a graphical model with nodes $I$ and edges $E$ parametrized by
\begin{align*}
    \textstyle
p(x \g \alpha, \beta) = \frac{1}{Z(\alpha, \beta)}\left( \sum_{i} \alpha_i + \sum_{i \ne j \in E} \beta_{ij} x_i x_j\right).
\end{align*}
For our Ising model, we use a $10\ttimes30$ lattice graph. In each experiment we set a strength parameter $\tau$, and randomly sampled each $\alpha_i$ and $\beta_{ij}$ from $\text{Uniform}[-\tau, \tau]$.

We estimated the standard deviations of the $1/Z$ estimators by computing the empirical root mean square error from the true value. It is possible to compute $Z(\alpha, \beta)$ exactly for the narrow strip we used. We also evaluated the empirical fraction of positive estimates for each algorithm.

Our importance sampling estimates were based on AIS \citep{neal2001annealed} using 10 intermediate distributions. We used the averaging scheme described in Section \ref{sec:avgscheme} and average over 10 AIS weights before taking reciprocals, which significantly improved variance. For our distribution on $N$, we choose the distribution satisfying $\Pr(N \ge k) \propto k^{-1.1}$.

Figure \ref{fig:varpos} shows the Markov-chain based estimators have lower variance and more positive estimates than IAE for lower values of $\tau$, where the importance samplers work well. We show the variance of the inverse importance weights as a reference to show how much debiasing increases variance. The variance of all three estimators increases as the importance sampling estimates become less reliable, but FCE degrades fastest because the Markov chain within FCE is more likely to ``stick''. However, FCE retains a significantly higher percentage of positive estimates, as expected theoretically.

At higher values of $\tau$, where the importance sampling estimates are less reliable, the IAE and RBBCE curves begin to look more similar. RBBCE still outperforms IAE in both variance and percent positive estimates for almost all values of $\tau$. In practice, however, it would make sense to improve the importance sampling estimates by increasing the number of intermediate annealing distributions and averaging over more estimates before applying debiasing schemes. In the setting where importance sampling estimates are already reliable, our Markov chain based estimators perform much better.

\begin{figure}[t]
\centering
\begin{subfigure}[b]{0.5\textwidth}
\centering
\includegraphics[scale=0.12]{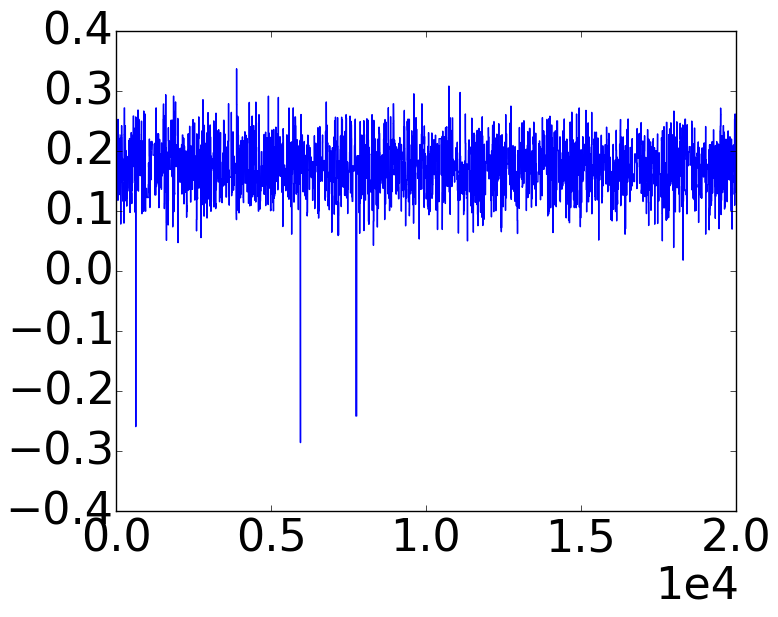}
\includegraphics[scale=0.12]{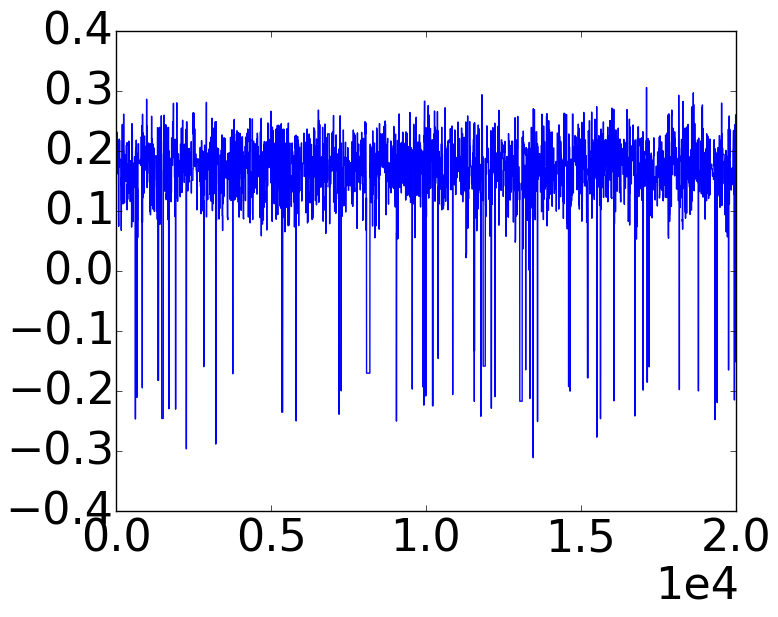}
\includegraphics[scale=0.12]{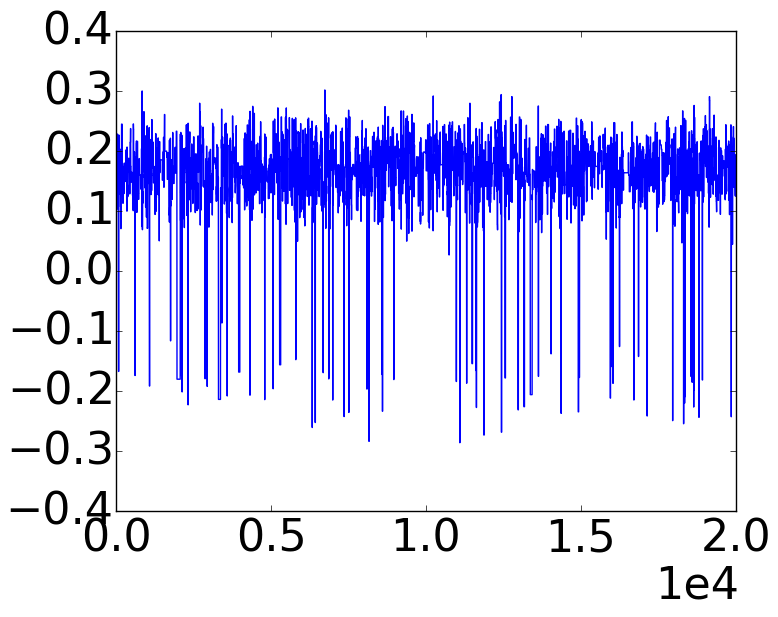}
\caption{Trace plots for $\sigma\beta$. From left to right: RBBCE, FCE, IAE}
\label{fig:betatraceising}
\end{subfigure}
\begin{subfigure} [b]{0.5\textwidth}
\centering
\includegraphics[scale=0.12]{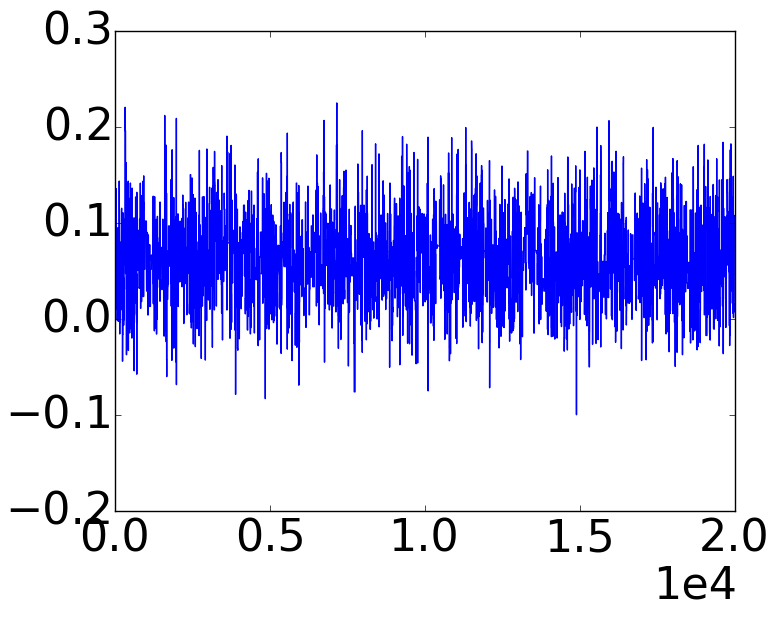}
\includegraphics[scale=0.12]{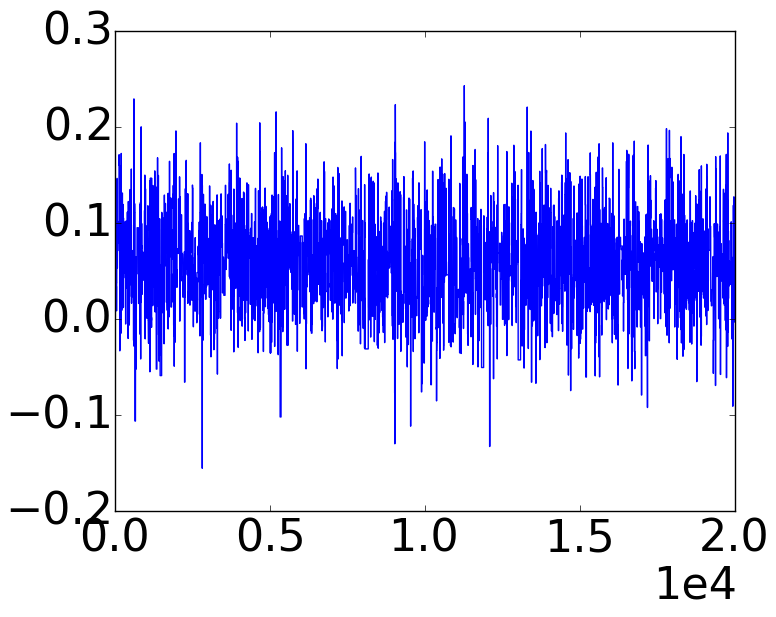}
\includegraphics[scale=0.12]{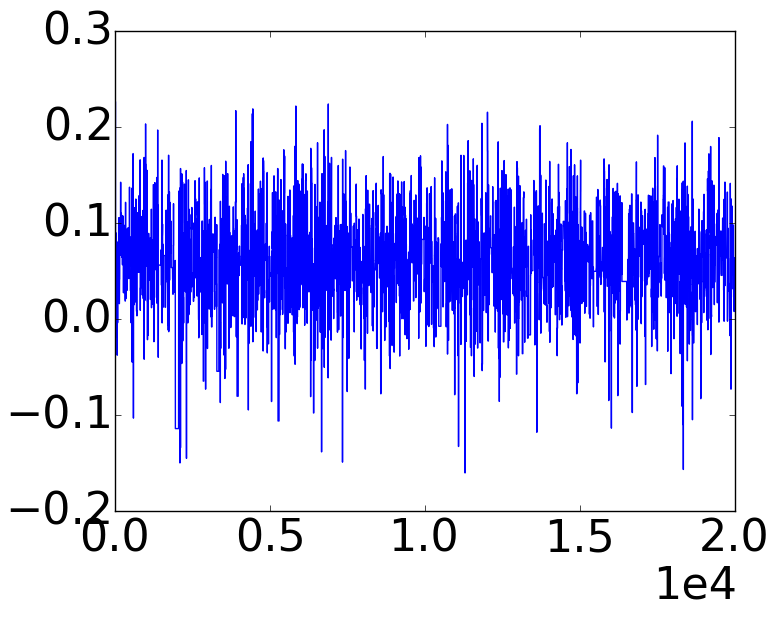}
\caption{Trace plots for $\sigma\alpha$. From left to right: RBBCE, FCE, IAE}
\label{fig:alphatraceising}
\end{subfigure}
\begin{subfigure} [b]{0.5\textwidth}
\centering
\includegraphics[scale=0.20]{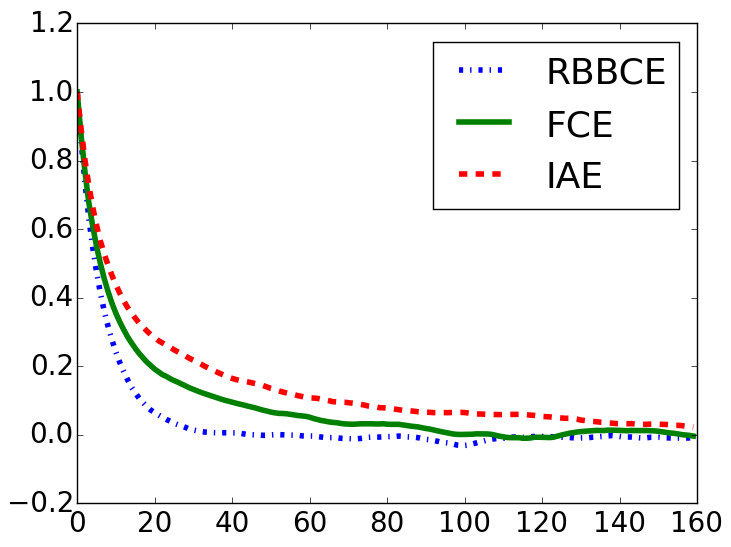}
\includegraphics[scale=0.20]{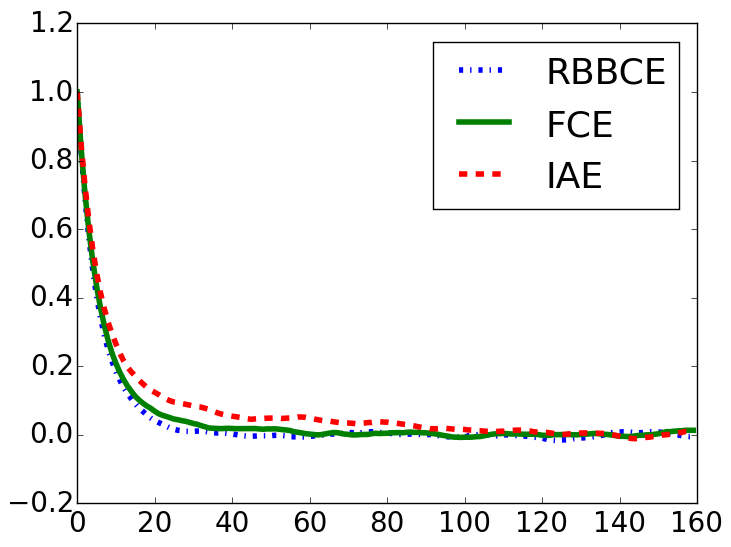}
\caption{Autocorrelation vs.\ lag for $\sigma\beta$ (left) and $\sigma\alpha$ (right).}
\label{fig:isingac}
\end{subfigure}
\caption{Trace and autocorrelation plots for doubly-intractable Ising runs.
All plots tracked parameters multiplied by $\sigma$, the sign of the estimator for $1/Z$. The autocorrelations without the sign term are roughly the same for all methods. Negative values in \subref{fig:betatraceising} result from negative $\sigma$, which gives high variance estimates.}
\label{fig:isingdi}
\vspace*{-0.2cm}
\end{figure}

\vspace*{-0.05cm}
\subsection{Pseudo-marginal Ising Grid}
\vspace*{-0.1cm}
We next tested the Russian Roulette algorithms in a pseudo-marginal estimation setting. We again run our experiments on a $10\ttimes30$ Ising lattice. We use a single bias and coupling parameter: $\alpha_i\te\alpha$ and $\beta_{ij}\te\beta$. Following \cite{murray2016}, we use uniform priors over $\alpha \in [-1, 1]$ and $\beta \in [0, 0.4]$. We used data generated with $\alpha \te 0.1$ and $\beta \te 0.1$. The pseudo-marginal Metropolis--Hastings outer loop used Gaussian proposals: $\alpha'\sim\N(\alpha,0.025^2)$ and $\beta'\sim\N(\beta,0.01^2)$, was run for 100,000 iterations, and used the method of \citet{lyne2015russian} for dealing with negative estimates. Our unbiased $1/Z$ estimator was averaged over 2 trials, each trial used weights formed by averaging 10 AIS weights with 30 intermediate distributions.

Figure \ref{fig:isingdi} shows the empirical autocorrelations and trace plots for our experiments. As $\beta\!>\!0$ negative values in the trace plot indicates a negative estimate for $1/Z$. Overall, out of 100,000 iterations, RBBCE had 99,924 positive samples while FCE and IAE had 97,597 and 96,538, respectively. As discussed by \citet{lyne2015russian}, a large fraction of positive estimates gives lower variance estimates of the posterior, which favours RBBCE\@.

We have not compared to the exchange algorithm \citep{murray2006}, which applies to this specific Ising model example. A direct comparison would be difficult: unlike our methods, the exchange algorithm depends on exact sampling, which has highly variable cost and depends on several additional details.

\begin{figure}[t]
\centering
\includegraphics[scale=0.21]{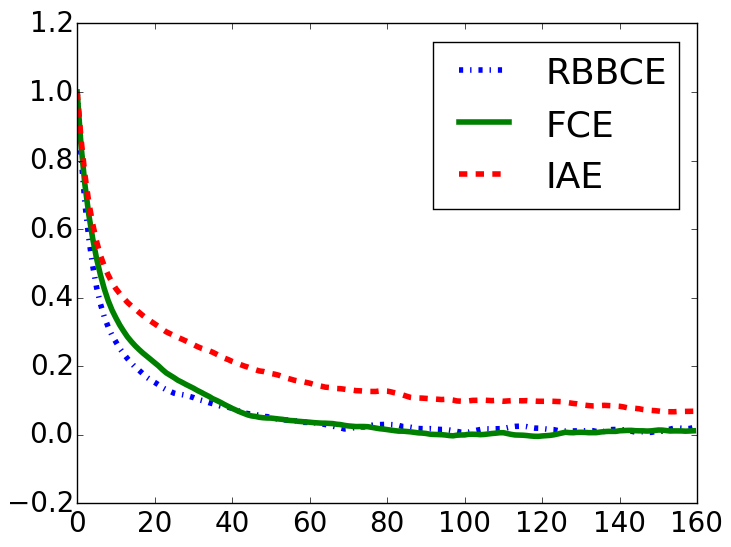}
\includegraphics[scale=0.21]{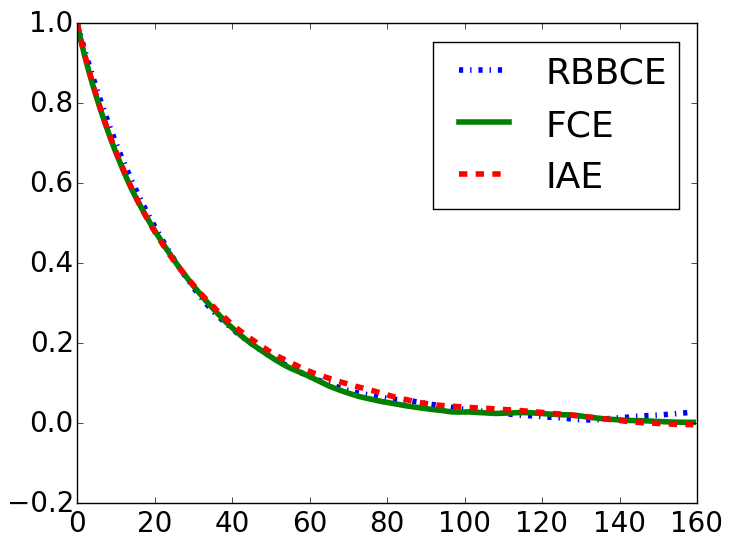}
\caption{Autocorrelation vs.\ lag for $\theta_e$ (left) and $\theta_s$ (right). As in Fig.~\ref{fig:isingac}, we plot signed autocorrelations. Markov chain based estimators exhibit less sticking.}
\label{fig:ergmdi}
\vspace*{-0.2cm}
\end{figure}

\vspace*{-0.1cm}
\subsection{Exponential Random Graph Model}
\vspace*{-0.2cm}
In our final demonstration, we apply the pseudo-marginal chains to Bayesian inference on exponential random graph models \citep{caimo2011bayesian}. These models capture relationships between sets of nodes, such as social interactions between individuals or formation of chemical structures between atoms. The distribution over graphs is\\

\vspace*{-1.1cm}
\begin{align*}
p(x \g \theta) = \exp(\theta^T s(x)) / Z(\theta),
\end{align*}
where $\theta$ are parameters and $s$ is a vector of sufficient statistics of
the graph $x$. \citet{caimo2011bayesian} used the exchange algorithm
\citep{murray2006} with approximate rather than exact samples. We believe our
experiments are the first application of an asymptotically correct MCMC method
to these models.

Our experiments use the Florentine graph, a social network graph modeling business relations between families in Florence in 1430. We let $\theta = (\theta_e, \theta_s)$, and $s = (\text{number of edges}, \text{average number of 2-stars per node})$, where a node with degree $d$ is involved in ${d \choose 2}$ 2-stars. We use a uniform prior for $\theta_e$ on $[-2.5, 2.5]$ and a uniform prior for $\theta_s$ on $[-1, 1]$. We run our pseudo-marginal chains for 100,000 iterations, averaging over 10 trials for each unbiased $1/Z$ estimator and using averages of 10 AIS weights with 10 intermediate distributions for our importance sampler. We tune Gaussian steps to $1$ for $\theta_e$ and $0.1$ for $\theta_s$.

Figure \ref{fig:ergmdi} shows the empirical autocorrelations of our chains. We report 99,890 positive estimates for RBBCE, 98,680 for FCE, and 98,442 for IAE\@. Although the improvements in positive estimates are more modest this time, our Markov chain based estimators still demonstrate lower autocorrelations than IAE\@.

\vspace*{-0.2cm}
\section{Discussion}
\vspace*{-0.2cm}
We introduced two novel algorithms, FCE and RBBCE, for producing unbiased estimates of $1/Z$ given access to black-box estimates unbiased in~$Z$. Our algorithms are generic, simple to implement, and perform debiasing at virtually no added cost. We are able to provide theoretical guarantees of finite expectation for many choices of state space (all choices for RBBCE) that hold regardless of the underlying distribution on truncation time. Unlike existing methods, these results allow valid use of the algorithms without needing to tune free parameters such as the growth rate of number of importance samples with truncation time.

FCE and RBBCE rely on Markov chain ``coupling'' with the motivation of improving variance and percentage of positive estimates, two heuristic indicators for how well our estimators would perform in a pseudo-marginal outer loop. Our experiments demonstrate that our algorithms can provide promising improvements over a non-coupling based debiasing scheme.

Our debiasing framework could be freely combined with recent developments in pseudo-marginal MCMC\@. For example \citet{doucet2015efficient}'s analysis could be used to tune the number of samples used for the $1/Z$ estimate. We could also apply pseudo-marginal slice sampling \citep{murray2016} with our algorithms.

\bibliography{refs}
\bibliographystyle{abbrvnat}
\end{document}